\documentclass[runningheads,a4paper]{llncs}

\usepackage{amssymb}
\usepackage{graphicx}
\usepackage{multirow}
\usepackage[draft]{todonotes}
\usepackage{xcolor}
\usepackage{lipsum} 
\usepackage{url}
\usepackage{color} 

\newcommand{\keywords}[1]{\par\addvspace\baselineskip
\noindent\keywordname\enspace\ignorespaces#1}

\begin{document}

\mainmatter  

\title{SLSDeep: Skin Lesion Segmentation Based on Dilated Residual and Pyramid Pooling Networks}

\titlerunning{Md. Mostafa Kamal Sarker et al.}

%
%

%
\author{Md. Mostafa Kamal Sarker\inst{1,\thanks{Corresponding Author: mdmostafakamal.sarker@urv.cat}} \and Hatem A. Rashwan\inst{1}  \and Farhan Akram\inst{2} \and Syeda Furruka Banu\inst{3} \and Adel Saleh\inst{1}  \and Vivek Kumar Singh\inst{1} \and Forhad U H Chowdhury \inst{4}   \and Saddam Abdulwahab\inst{1}\and Santiago Romani\inst{1} \and Petia Radeva\inst{5} \and Domenec Puig\inst{1}} 

\authorrunning{Md. Mostafa Kamal Sarker et al.}


\institute{Department of Computer Engineering and Mathematics, Universitat Rovira i Virgili, 43007 Tarragona, Spain. \and
Imaging Informatics Division, Bioinformatics Institute, 30 Biopolis Street, $\#$ 07-01 Matrix, 138671, Singapore.
\and Department of Technology and Engineering Management, Rovira i Virgili University, 43007 Tarragona, Spain.
\and Liverpool School of Tropical Medicine, Liverpool L3 5QA, UK. \and
Department of Mathematics, University of Barcelona, 08007 Barcelona, Spain.\\
}
%

\toctitle{Md. Mostafa Kamal Sarker et al.}
\maketitle

\begin{abstract}
Skin lesion segmentation (SLS) in dermoscopic images is a crucial task for automated diagnosis of melanoma. In this paper, we present a robust deep learning SLS model, so-called SLSDeep, which is represented as an encoder-decoder network. The encoder network is constructed by dilated residual layers, in turn, a pyramid pooling network followed by three convolution layers is  used for the decoder. Unlike the traditional methods employing a cross-entropy loss, we investigated a loss function by combining both Negative Log Likelihood (NLL) and End Point Error (EPE) to accurately segment the melanoma regions with sharp boundaries. The robustness of the proposed model was evaluated on two public databases: ISBI 2016 and 2017 for skin lesion analysis towards melanoma detection challenge. The proposed model outperforms the state-of-the-art methods in terms of segmentation accuracy. Moreover, it is capable to segment more than $100$ images of size $384\times 384$ per second on a recent GPU.

\end{abstract}

\section{Introduction}

According to the skin Cancer Foundation Statistics, the percentage of both melanoma and non-melanoma skin cancers has been increasing rapidly over the last few years~\cite{CAAC:CAAC21387}. Dermoscopy, non-invasive dermatology imaging methods, can help the specialists to inspect the pigmented skin lesions and diagnose malignant melanoma at an initial-stage \cite{kardynal2014modern}. Even the professional dermatologist can not properly classify the melanoma only by relying on their perception and vision. Sometimes, human tiredness and other distractions during visual diagnosis can also yield high number of false positives\cite{celebi2008automatic}. Therefore, a Computer-Aided Decision system (CAD)  is needed to assist the dermatologists to properly analyze the dermoscopic images and accurately segment the melanomas. Many attempts of melanoma segmentation have been proposed in the literature. An overview of the different melanoma segmentation techniques is presented in {\cite{zhang2017melanoma}}. However, this task is still a challenge, since the dermoscopic images has various characteristics including different sizes and shapes, fuzzy boundaries, different colors, and the presence of hair \cite{day2000automated}.
\begin{figure}[htp]
\centering
\includegraphics[width=\textwidth]{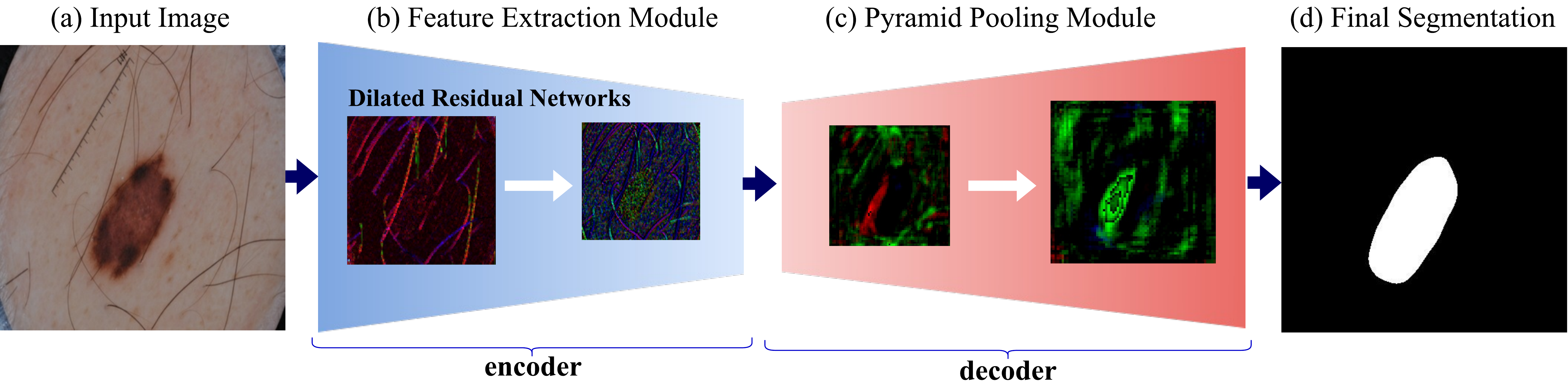}
\caption{Architecture of the proposed skin lesion segmentation network.}
\label{fig1:model}Negative Log Likelihood (NLL) and End Point Error (EPE)
\end{figure}
In last few decades, many approaches have been proposed to cope with the aforementioned challenges. Most of these methods are based on thresholding, edge-based and/or region-based active contour models, clustering and supervised learning \cite{celebi2015state}. However, these methods are unreliable when dermoscopic images are inhomogeneous and/or lesions have fuzzy or blurred boundaries\cite{celebi2015state}. Furthermore, their performance relies on efficient pre-processing algorithms, such as filtering, illumination correction and hair removal, which badly affect the generalizability of these models.

Recently, deep learning methods applied to image analysis, specially Convolutional Neural Networks (CNNs) have been used to solve the image segmentation problem\cite{long2015fully}. These CNN-based methods can automatically learn features from raw pixels to distinguish between background and foreground objects to attain the final segmentation. Most of these approaches generally are based on encoder-decoder networks~\cite{long2015fully}. These networks learn to map the features of an image to a segmented image. The encoder networks are used for extracting the features from the input images, in turn the decoder ones used to construct the segmented image. The U-net network proposed in~\cite{ronneberger2015u} has been particularly designed for biomedical image segmentation based on the concept of Fully Convolutional Networks(FCN)~\cite{long2015fully}. The U-net model reuses the feature maps of the encoder layers to the corresponding decoders and concatenates them to upsampled (via deconvolution) decoder feature maps called ``skip-connections''. The U-Net model for SLS outperformed many classical clustering techniques~\cite{lin2017skin}. 

In addition, the deep residual network (ResNet) model~\cite{yu2017automated} is a 50-layers network designed for segmentation tasks. ResNet blocks are used to boost the overall depth of the networks and allow more accurate segmentation depending on more significant image features. Moreover, Dilated Residual Networks (DRNs) proposed in~\cite{yu2017dilated} increase the resolution of the ResNet blocks’s output by replacing a subset of interior subsampling layers by dilation~\cite{yu2015multi}. DRNs outperform the normal ResNet without adding algorithmic complexity to the model. DRNs are able to represent both tiny and large image features. Furthermore, Zhao et. al.~\cite{zhao2017pyramid} proposed a Pyramid Pooling Network (PPN) that is able to extract additional contextual information based on a multi-scale scheme. 
  
Inspired by the success of the aforementioned deep models for semantic segmentation, we propose a model combining skip-connections, dilated residual and pyramid pooling networks for SLS with different  improvements. In our model, the encoder network depends on DRNs layers, in turn the decoder depends on a PPN layer along with their corresponding connecting layers. More features can be extracted from the input dermoscopic images by combining DRNs with PPN, in turn it also enhances the performance of the final network. Finally, our SLS segmentation model uses a new loss function, which combines Negative Log Likelihood (NLL) and End Point Error (EPE)~\cite{baker2011database}. Mainly, cross-entropy is used for multi-class segmentation models, however it is not as useful as NLL in binary class segmentation. Thus, in such melanoma segmentation, we propose to use NLL as a loss function. In addition, for preserving the melanoma boundaries, EPE is used as a content loss function. Consequently, this paper aims at developing an automated deep SLS model with two main contributions:
\begin{itemize}
  \item [$\bullet$] An encoder-decoder network for efficient SLS without any pre- and post-processing algorithms based on dilated residual and pyramid pooling networks to enclose coarse-to-fine features of dermoscopic images.
  \item [$\bullet$] A new loss function that is a combination of Negative Log Likelihood and End Point Error for properly detecting the melanoma with sharp edges.
\end{itemize}

\section{Proposed Model}
\subsection{Network Architecture}
\begin{figure}[htp]
\centering
\includegraphics[width=\textwidth]{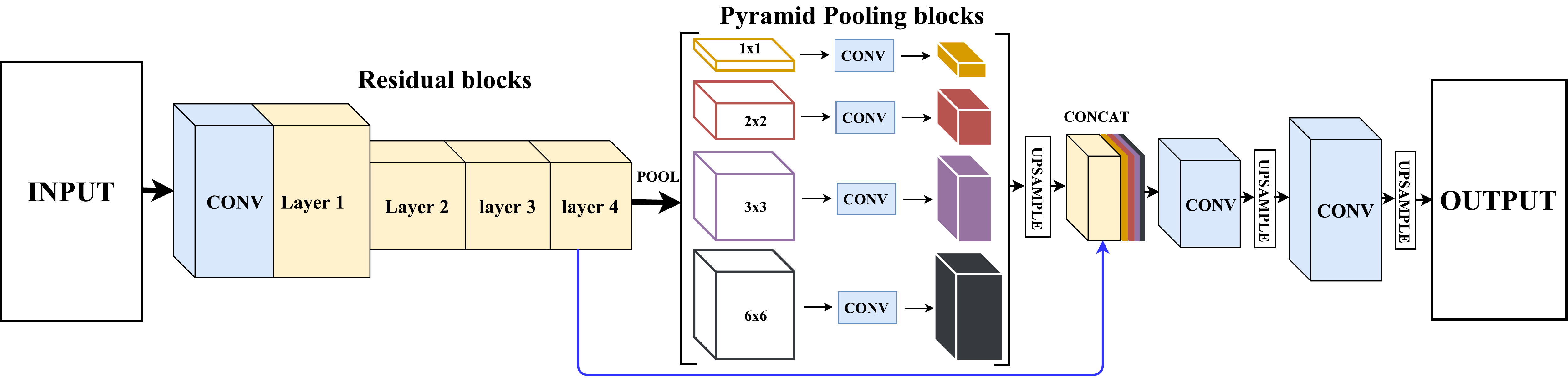}
\caption{Architecture of the Encoder-decoder network.}
\label{fig2:model2}
\end{figure}
Fig.\ref{fig1:model} shows the architecture of the proposed SLSDeep model with DRNs~\cite{zhou2017scene} and PPN~\cite{he2015spatial}. The network contains two-fold architecture: encoder and decoder. Regarding the encoder phase, the first layer is a $3\times 3$ convolutional layer followed by $3\times 3$ max pooling with stride 2.0 that generates 64 feature maps. This layer uses Relu as an activation and batch normalization to speed-up the training steps with a random initialization. Following, four pre-trained DRNs blocks are then used to extract 256, 512, 1024 and 2048 feature maps, respectively as shown in Fig.\ref{fig2:model2}. The first, third, and fourth DRNs layers are with stride 1.0, in turn the second one is with stride 2.0. Thus, the size of final output of encoder is $1/8$ of the input image (e.g. in our model, the input image is in $384\times 384$ and the output feature maps of the encoder is $48\times 48$). For global contextual prior, average pooling is used before feeding to fully connected layers in image classification \cite{szegedy2015going}. However, it is not sufficient to extract necessary information from our skin lesion images. Therefore, we do not use average pooling at the end of the encoder and directly fed the output feature maps to the decoder network. 

On the other hand, for the decoder network, we use the concept of PPN for producing multi-scale (coarse-to-fine) feature maps and then all scales are concatenated together to get more robust feature maps. PPN use a hierarchical global prior of variant size feature maps in multi-scales with different spatial filters as shown in Fig.\ref{fig2:model2}. In this paper, the used PPN layer extracts feature maps using four pyramid scales with rescaling sizes of $ 1 \times 1, 2 \times 2, 3 \times 3 $ and $6 \times 6$. A convolutional layer with a $1 \times 1$ kernel in every pyramid level is used for generating 1024 feature maps. The low-dimension feature maps are then upsampled based on bilinear interpolation to get the same size of the input feature maps. The input and four feature maps are finally concatenated to produce 6144 feature maps (i.e., 4x1024 feature maps concatenated with the input 2048 feature maps). Sequentially, two $3\times 3$ convolutional layers are followed by two upsampling layers. Finally, a softmax function (i.e. normalized exponential function) is utilized as logistic function for producing the final segmentation map. A ReLU activation with batch normalization is used in the two convolutional layers ~\cite{ioffe2015batch}. Moreover, in order to avoid the overfitting problem, the dropout function with a ratio of 0.5~\cite{srivastava2014dropout} is used before the second upsampling layer. 

The skip connections between all layers of the encoder and decoder were tested during the experiments. However, the best results were provided when only one skip connection was done between the last layer of the encoder and the output of PPN layer of the decoder. The architecture of the encoder and decoder is given in details with the supplementary materials.

\subsection{Loss Function}

Most of the traditional deep learning methods commonly employ cross-entropy as a loss function for segmentation~\cite{ronneberger2015u}. Since the melanoma is mostly a small part of a dermoscopic image, the minimization of cross-entropy tends to be biased towards the background. To cope with this challenge, we propose a new loss function by combining objective and content losses: NLL and EPE, respectively. In order to fit a log linear probability model to a set of binary labeled classes, the NLL is the objective function of the proposed model to minimize. Let $ v \in \{0,1\}$ be a true label for binary classification and $ p = {Pr}(v = 1)$ a probability estimate, the NLL of the binary classifier can be defined as:
\begin{equation}
	L_{log}(v,p) = - \log Pr(v|p)= -(v\log(p)+(1-v)\log(1-p)).
	\label{eq:logloss1}
\end{equation}
Regarding the content of the loss function, we have also computed an absolute error aiming at maximizing the Peak Signal-to-Noise Ratio (PSNR) by preserving the object boundaries. The used EPE loss follows a classical approach that the generated mask, $t{u}$ is pixel-wise compared with the corresponding ground-truth, $v$. The EPE error can be defined by \cite{baker2011database}:
\begin{equation}
	L_{epe}= \sqrt{({u}_{0}-u_{1})^2+({v}_{0}-v_{1})^2}
	\label{eq:epeloss}
\end{equation}
where ${u}_0$ and ${u}_1$ are the first derivatives of ${u}$ in $x$ and $y$ directions, and $v_0$ and $v_1$ are the first derivatives of $v$.

Hence, our final loss, which combines both NLL and EPE, can be defined as:
\begin{equation}
L_{total}=L_{log}+ \alpha L_{epe}
	\label{eq:loss}
\end{equation}
where $\alpha < 1$ is a weighted coefficient. In this work, we use $\alpha=0.5$.

\section{Experimental Setup and Evaluation} \
\textbf{Database:} To test the robustness of the proposed model, it was evaluated on two public benchmark datasets of dermoscopy images for skin lesion analysis: \textbf{ISBI 2016}~\cite{codella2017skin} and \textbf{ISBI 2017}~\cite{gutman2016skin}. The datasets images are captured by different devices at various top clinical centers over the world. In ISBI 2016 dataset, training and testing part contain 900 and 379 annotated images, respectively. The size of the images ranges from $542\times 718$ to $2848\times 4288$ pixels. In turn, ISBI 2017 dataset is divided into training, validation and testing parts with 2000, 150 and 600 images, respectively. 

\textbf{Evaluation Metrics:} We used the evaluation metrics of ISBI 2016 and 2017 challenges for evaluating the segmentation performances including Specificity(SPE), Sensitivity(SEN), Jaccard index(JAC), Dice coefficient(DIC) and Accuracy(ACC) detailed in \cite{gutman2016skin} and \cite{codella2017skin}. 

\textbf{Implementation:} The proposed model is implemented on an open source deep learning library named PyTorch\cite{paszke2017pytorch}. For optimization algorithm, we used Adam \cite{kingma2014adam} for adjusting learning rate, which depends on first and second order moments of the gradient. We used a ``poly'' learning rate policy \cite{chen2016deeplab} and selected a base learning rate of 0.001 and 0.01 for encoder and decoder, respectively with a power of 0.9. For data augmentation, we selected random scale between 0.5 and 1.5, random rotation between -10 and 10 degrees. The ``batchsize'' is set to 16 for training and the epochs to 100. The experiments utilized NVIDIA TITAN X with 12GB memory and its takes approximately 20 hours for train the networks.

\textbf{Evaluation and results:} Since the size of the given images is very large, we resized the input images into $384\times 384$ pixels for training our model. In this work, we tested different sizes and the $384\times 384$ size yields the best results. In order to separately assess the different contributions of this model, the resulting segmentation for the proposed model with different variations have been computed: (a) The SLSDeep model without the content loss EPE (SLSDeep-EPE), (b) the proposed method with skip connections of all encoder and decoder layers (SLSDeep+ASC) and (c) the final proposed model (SLSDeep) with NLL and EPE loss functions and only one skip connection between the last layer of the encoder and the PPN layer.

Quantitative results on ISBI'2016 and ISBI'2017 datasets are shown in Table~\ref{table2}. Regarding ISBI'2016, we compared the SLSDeep and its variations to the four top methods: ExB, \cite{yu2017automated}, \cite{rahman2016developing} and \cite{yuan2017automatic} providing the best results according to \cite{gutman2016skin}. The segmentation results of our model SLSDeep with its variations (SLSDeep-EPE and SLSDeep+ASC) perform much better than the all evaluated methods on the ISBI'2016 with the five aforementioned evaluation metrics. SLSDeep yields the best results among the three variations. In addition, for the DIC score, our model, SLSDeep, improved the results with around $3.5\%$, while the JAC score was significantly improved with $8\%$. The SLSDeep yielded results with an overall accuracy of more than $98\%$. 

\begin{table}[!ht]
	\centering
	\caption{Performance Evaluation on the ISBI Challenges Dataset}
	\label{table2}
	\resizebox{\textwidth}{!}{%
		\smallskip
		\begin{tabular}{c|lllllll}
			\hline
			Challenges          &\quad Methods                                &\quad ACC   &\quad DIC   &\quad JAC   &\quad SEN   &\quad SPE   \\ \hline\hline
			\multirow{4}{*}{ISBI 2016}    & ExB                                &\quad 0.953 &\quad 0.910 &\quad 0.843 &\quad 0.910 &\quad 0.965 \\ 
			& CUMED\cite{yu2017automated}                 &\quad 0.949 &\quad 0.897 &\quad 0.829 &\quad 0.911 &\quad 0.957 \\  
			& Rahman et. al.\cite{rahman2016developing}   &\quad 0.952 &\quad 0.895 &\quad 0.822 &\quad 0.880 &\quad 0.969 \\
			& Yuan et. al.\cite{yuan2017automatic}        &\quad 0.955 &\quad 0.912 &\quad 0.847 &\quad 0.918 &\quad 0.966 \\           
			& \textbf {SLSDeep}                           &\quad \textbf{0.984} &\quad\textbf{0.955} &\quad\textbf{0.913} &\quad\textbf{0.945} &\quad\textbf{0.992} \\ 
			& \textbf {SLSDeep-EPE}                       &\quad 0.973 &\quad 0.919 &\quad 0.850 &\quad 0.890 &\quad 0.990 \\ 
			& \textbf {SLSDeep+ASC}                       &\quad 0.975 &\quad 0.930 &\quad 0.869 &\quad 0.952 &\quad 0.979 \\ \hline
			\multirow{4}{*}{ISBI 2017}   & Yuan et. al.\cite{yuan2017automatic}    &\quad 0.934 &\quad 0.849 &\quad 0.765 &\quad\textbf{0.825}  &\quad 0.975 \\  
			& Berseth et. al.\cite{berseth2017isic} &\quad 0.932 &\quad 0.847 &\quad 0.762 &\quad 0.820 &\quad 0.978 \\  
			& MResNet-Seg\cite{bi2017automatic}     &\quad 0.934 &\quad 0.844 &\quad 0.760 &\quad 0.802 &\quad 0.985 \\ 
			& \textbf {SLSDeep}         &\quad \textbf{0.936}       &\quad\textbf{0.878}        &\quad\textbf{0.782}      &\quad0.816      &\quad 0.983    \\ 
			& \textbf {SLSDeep-EPE}         &\quad  0.913       &\quad 0.826        &\quad 0.704      &\quad 0.729     &\quad \textbf{0.986}   \\
			& \textbf {SLSDeep+ASC}         &\quad  0.906      &\quad 0.850       &\quad 0.739     &\quad 0.808      &\quad 0.905   \\\hline 
			
		\end{tabular}%
	}
\end{table}

Furthermore,  SLSDeep on the ISBI'2017 provided segmentation results with improvements of $3\%$ and $2\%$ in terms of DIC and JAC scores, respectively. Again, SLSDeep perform better the three top methods of the ISBI'2017 benchmark,~\cite{yuan2017automatic}, \cite{berseth2017isic} and \cite{bi2017automatic}, with ACC, DIC and JAC scores. However, \cite{yuan2017automatic} yielded the best SEN score with an improvement of $0.9\%$ better than our model. The SLSDeep-EPE and SLSDeep+ASC provided reasonable results, however their results were worse than the three tested methods. SLSDeep-EPE yields the highest SPE, which is $0.1\%$ and $0.3\%$ more than MResNet-Seg \cite{bi2017automatic} and SLSDeep, respectively. Using the EPE function with the final SLSDeep model significantly improved the DIC and JAC scores of $3\%$ and $5\%$, respectively, on ISBI'2016 and of $5\%$ and $8\%$, respectively, with ISBI'2017. In addition, SLSDeep with only one skip connections yields better results than SLSDeep+ASC on both ISBI datasets.

\begin{figure}[!ht]
	\centering
	\includegraphics[width=\textwidth]{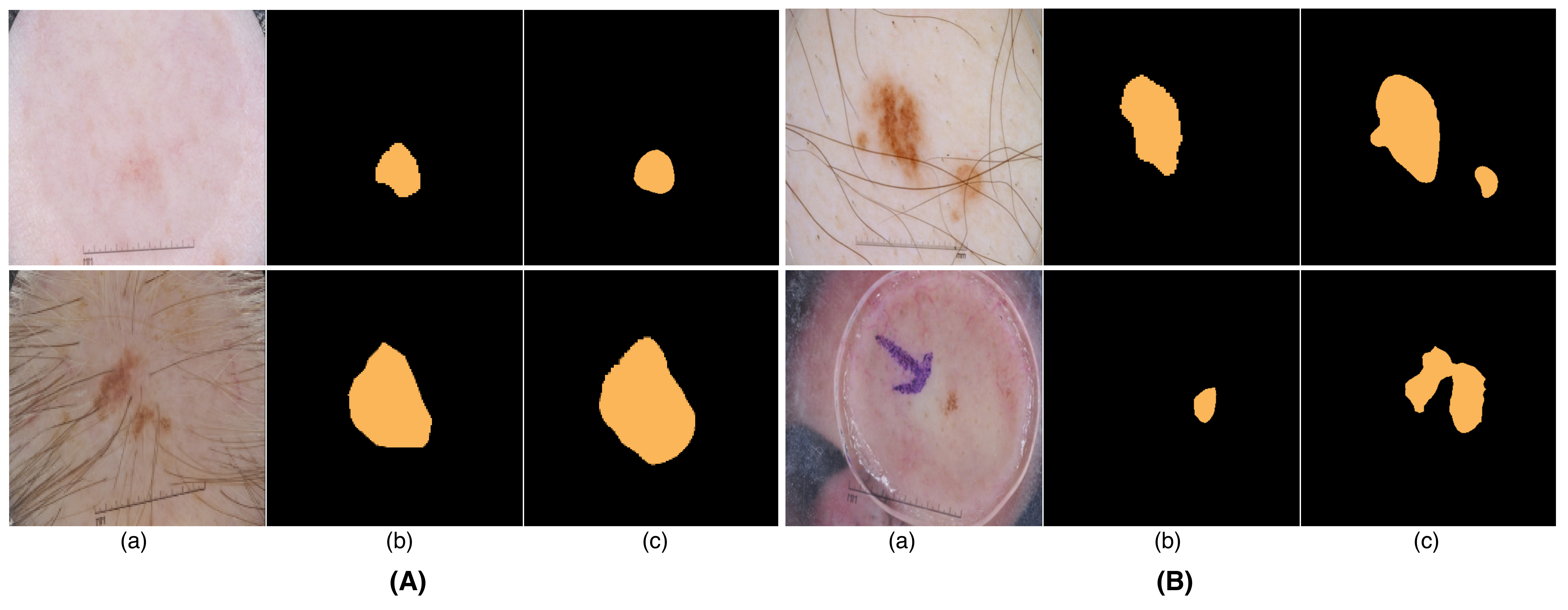}
	\caption{
		Segmentation results: \textit{(a) input image, (b) ground truth and (c) segmentation result}; (A) correct segmentation by our model; (B) incorrect segmentation by our model.}
	\label{fig3:results}
\end{figure}   

Qualitative results of four examples of the ISBI'2017 dataset are shown in Fig.\ref{fig3:results}. For the first and second examples (on the top-and down-left side), the lesions were properly detected, although the color of the lesion area is very similar to the rest of the skin. In addition, the lesion area was accurately segmented with sharp edges. Regarding to the third example (on the top-right side), SLSDeep properly segmented the lesion area; however a small false region with similar features was also detected. In turn, the last example is very difficult, since the lesion shown in the input image is a very small region. However, the SLSDeep model can segment it, but with bigger size of false negative regions.

\section{Conclusions}
This paper proposed a novel deep learning skin lesion segmentation model based on training an encoder-decoder network. The encoder network used the dilated ResNet layers with downsampling to extract the features of the input image, in turn convolutional layers with pyramid pooling and upsampling are used to reconstruct the segmented image. This approach outperforms, in terms of skin lesion segmentation, the literature evaluated on two ISBI'2016 and ISBI'2017 datasets. The experiments show that SLSDeep is robust segmentation technique using different evaluation metrics: accuracy, Dice coefficient, Jaccard index and specificity. In addition, the qualitative results show promising skin lesion segmentation. For future work, the proposed model will be explored on different color spaces  and applied to other medical applications to prove its versatility.

\bibliographystyle{splncs03}
\bibliography{my_ref}

\end{document}